\documentclass[12pt]{article}
\usepackage{amssymb,amsmath,makeidx, amsthm}
\begin{document}
\def\ol{\overline}
\def\wt{\widetilde}
\def\wh{\widehat}
\def\mn{\medskip\noindent}
\def\n{\noindent}

\title{BACKPROPAGATION IN MATRIX NOTATION\\
{\small To Student and Lecturer relaxation}}
\author{ N. M. Mishachev
  \\ Lipetsk Technical University, \\
     Lipetsk, 398055 Russia\\
      {\small nmish@lipetsk.ru}}

\date{}
\maketitle

\begin{abstract}

\n
In this note we calculate the gradient of the network function in
matrix notation.  
\end{abstract}

\mn
{\large\bf 1.  Introduction}

\mn
A feed-forward neural network 
is a composition of alternating linear and coordinate-wise non-linear maps,  
where the matrices of linear maps are considered 
as adjustable (=optimizable) network parameters.
Unlike the conventional regression models, the adjustable parameters enter 
the neural network non-linearly.
To optimize (= to train) the neural network, we need to
calculate, using the chain rule, the gradient of the network
with respect to parameters.
Formulas for the gradient were obtained by a number of authors
(it is customary to refer to [WHF]), and the corresponding
calculation were called the {\it backpropagation} algorithm.
In the recursive-coordinate form we can find these formulas in any 
neural networks manual. Less often in these tutorials one can find an
explicit (non-recursive) matrix version of the formulas, although the 
matrix representation for the neural network itself usually appears in books.
Meanwhile, looking at the matrix representation for
neural network, everyone (I think) wants to see
in explicit matrix form both the final formula
and entire calculation.
In any case, when I began to teach students this subject,
I was annoyed by the number of indices in the usual coordinate-recursive 
proof. To get rid of the indexes, I tried to apply the chain rule 
directly in matrix form. This undergraduate-level exercise 
turned out to be not too simple.
Looking back at the result (see the text), 
I am not sure that the matrix-style proof has became simpler then in
recursive-coordinate form. Still I hope that the text can be of some use. 

\mn
{\large\bf 2. Network functions}

\mn
Everywhere further the vectors $X\in\mathbb R^n$ are considered as columns
$X=[x_1,\dots,x_n]^T$. The component-wise product (= Hadamard product) 
of matrices of the same dimension is denoted by $A\circ B$.
The coordinate-wise map $\Sigma:\mathbb R^n\to \mathbb R^n$,
defined by the formula
\begin{equation*}
\Sigma(X)=\Sigma([x_1,\dots,x_n]^T)=[\sigma_1(x_1),\dots,\sigma_k(x_n)]^T,
\end{equation*}
i.e. the direct sum of $n$ functions $\sigma_i:\mathbb R^1\to\mathbb R^1$,
is denoted by double arrow $\Sigma:\mathbb R^n\rightrightarrows\mathbb R^n$.
The action of such a map on the vector $X$ can be considered as an
``operator" Hadamard product of columns $\Sigma=[\sigma_1,\dots,\sigma_k]^T$
and $X=[x_1,\dots,x_n]^T$, i.e.  $\Sigma(X)=\Sigma\circ X$.
{\it A (neural) network function} $f:\mathbb R^n\to \mathbb R^1$ is a function of the form
\begin{equation*}
f:\mathbb R^n\xrightarrow{W_1} \mathbb R^{n_1}\mathop
\rightrightarrows^{\Sigma_1}\mathbb R^{n_1}
\xrightarrow{W_2}\mathbb R^{n_2}\mathop\rightrightarrows^{\Sigma_2}\mathbb R^{n_2}
\xrightarrow{W_3}\dots
\xrightarrow{W_{k-1}}\mathbb R^{n_{k-1}}
\mathop\rightrightarrows^{\Sigma_{k-1}}\mathbb R^{n_{k-1}}
\xrightarrow{W_k} \mathbb R^{1}\mathop\rightarrow^{\Sigma_k}\mathbb R^{1} 
\end{equation*}
where $W_i$ are linear maps and $\Sigma_i=[\sigma_{i1},\dots,\sigma_{i{n_i}}]^T$
are coordinate-wise maps.
Functions $\sigma_{ij}:\mathbb R^1\to\mathbb R^1$ are called
{\it activation functions}. Their choice is important in the neural networks theory;
here it will be enough to assume that $\sigma_{ij}$ are arbitrary 
piecewise differentiable functions.
Note that $\Sigma_k=[\sigma_{k1}]^T=\sigma_k$ as $n_k=1$.
The map $f$ can be written as
\begin{equation*}
f(X)=f(X;W)=\Sigma_k(W_k\cdot\Sigma_{k-1}(W_{k-1}\cdot\Sigma_{k-2}
\dots\cdot\Sigma_2(W_2\cdot\Sigma_1(W_1\cdot X))\dots))\,
\end{equation*}
or, alternatively, as the product 
\begin{equation*}
f(X)=f(X;W)=\Sigma_k\circ W_k\cdot\Sigma_{k-1}\circ W_{k-1}\cdot\Sigma_{k-2}
\dots\cdot\Sigma_2\circ W_2\cdot\Sigma_1\circ W_1\cdot X
\end{equation*}
(actions order is from the right to the left),
where $X=[x_1,\dots,x_n]^T\in \mathbb R^n$ and $W_i$ are matrices of the same 
name linear maps $W_i$. These matrices are considered to be parameters of 
the map $f(X;W)$, so the parameter vector $W=(W_k,\dots,W_1)$ 
consist of  $n_{i}\times n_{i-1}$ matrices $W_i$ ($n_k=1\,,\,\,n_0=n$).
We also set $N_1(X)=W_1\cdot X$ and $N_{i+1}(X)=W_{i+1}\cdot\Sigma_{i}(N_{i}(X))$,
then
\begin{equation*}
f(X;W)=\Sigma_k(\underbrace{W_k\cdot\Sigma_{k-1}(\underbrace{W_{k-1}\cdot\Sigma_{k-2}
\dots\cdot\Sigma_2(\underbrace{W_2\cdot\Sigma_1
(\underbrace{W_1\cdot X}_{N_1})}_{N_2})\dots}_{N_{k-1}})}_{N_k})
\end{equation*}
and
\begin{equation*}
f(X;W)=\Sigma_k(N_k)=\Sigma_k(W_k\cdot\Sigma_{k-1}(N_{k-1}))=\dots\,, 
\end{equation*}
where $N_k,\dots,N_1$ are the columns of dimensions $n_k,\dots,n_1$.
Note that for $k = 1$ and identity activation function $\Sigma_k=\sigma_1$
the network function is a homogeneous linear function of $n$ variables.

\mn
{\large\bf 3.  Network functions and neural networks}

\mn
The network function of the form (1) defines a {\it homogeneous} 
$k$-layered feed-forward neural network with $n$-dimensional input 
and one-dimensional output. The number of hidden (internal)
layers is $k-1$ and the $i$-th layer contains $n_i$ neurons.
The word {\it homogeneous} in this context
is not generally accepted and  
means that all intermediate linear maps are homogeneous.     
The conventional {\it affine} network, in which linear maps 
are supplemented with biases, can be obtained from the homogeneous one if: 
\begin{itemize}
\item[(a)] all input vectors have the form $X=[x_1,\dots,x_{n-1},1]^T$;
\item[(b)] the last rows of all matrices $W_i$ with $i<k$ have the form $[0,\dots,0,1]$;
\item[(c)] the last functions $\sigma_{in_i}$ in the columns $\Sigma_i$ 
with $i<k$ are identical. 
\end{itemize}
In this instance, the homogeneous network will be equivalent to a $k$-layered
affine neural network with $(n-1)$-dimensional input and one-dimensional 
output. Each hidden layer of such a network will contain $n_i-1$ ``genuine'' 
neurons and one (last) ``formal'', responsible for the bias;
the last column  of the matrix $W_i$, except the last element,
will be the bias vector for the $i$-th layer.
For $k = 1$ and identical activation function $\sigma_1=\rm id$ 
the described adjustment corresponds to the standard transition from a 
homogeneous multiple regression to a non-homogeneous
by adding to the set of independent variables (predictors) an
additional formal predictor, always equal to one.

\mn
{\large\bf 4.  Gradient $\nabla_Wf(X;W)$ in one-dimensional case}
 
\mn
Let $n=n_1=\dots=n_k=1$. Then all matrices $W_i$ are numbers $w_i$, all 
columns $\Sigma_i$ are functions $\sigma_i$ and
\begin{equation*}
f(x)=f(x;W)=\sigma_k(w_k\sigma_{k-1}(w_{k-1}\sigma_{k-2}(w_{k-2}\sigma_{k-3}
\dots\sigma_2(w_2\sigma_1(w_1 x))\dots)))\,.
\end{equation*}
The application of the chain rule does not cause 
difficulties, and for the gradient

\begin{equation*}
\nabla_W f=(\nabla_{w_k}f,\nabla_{w_{k-1}}f,\nabla_{w_{k-2}}f\dots ,\nabla_{w_1}f)
\end{equation*}
we obtain
\begin{equation}
\begin{cases}
\nabla_{w_k}f=\overbrace{\sigma_k'(N_k)\,\sigma_{k-1}(N_{k-1})}\\
\nabla_{w_{k-1}}f=\underbrace{\sigma_k'(N_k)\,w_k}\,
\overbrace{\sigma_{k-1}'(N_{k-1})\,\sigma_{k-2}(N_{k-2})}\\
\nabla_{w_{k-2}}f=\underbrace{\sigma_k'(N_k)\,w_k}\,
\underbrace{\sigma_{k-1}'(N_{k-1})\,w_{k-1}}\,
\overbrace{\sigma'_{k-2}(N_{k-2})\,\sigma_{k-3}(N_{k-3})}\\
...................................................................................................\\
\nabla_{w_1}f=\underbrace{\sigma_k'(N_k)\,w_k}\,
\underbrace{\sigma_{k-1}'(N_{k-1})\,w_{k-1}}\,
\underbrace{\sigma'_{k-2}(N_{k-2})\,w_{k-2}}\dots\,
\overbrace{\sigma'_{1}(N_{1})\,x}\\
\end{cases}
\end{equation}
or, omitting the arguments $N_i$  for brevity,
\begin{equation}
\begin{cases}
\nabla_{w_k}f=\overbrace{\sigma'_k\,\sigma_{k-1}}\\
\nabla_{w_{k-1}}f=\underbrace{\sigma'_k\,w_k}\,
\overbrace{\sigma_{k-1}'\,\sigma_{k-2}}\\
\nabla_{w_{k-2}}f=\underbrace{\sigma_k'\,w_k}\,
\underbrace{\sigma_{k-1}'\,w_{k-1}}\,
\overbrace{\sigma'_{k-2}\,\sigma_{k-3}}\\
..............................................................\\
\nabla_{w_1}f=\underbrace{\sigma_k'\,w_k}\,
\underbrace{\sigma_{k-1}'\,w_{k-1}}\,
\underbrace{\sigma'_{k-2}w_{k-2}}\,\dots\,
\underbrace{\sigma'_{2}w_{2}}\overbrace{\sigma'_{1}\,x}.\\
\end{cases}
\end{equation}
Here the braces emphasize  the periodicity in 
the structure of the formulas.
In order to write down the recursive formula we set, omitting again 
the arguments in the notations,
\begin{equation*}
\Delta_{i}=\Delta_{i+1}\,w_{i+1}\,\sigma_{i}'\,,
\end{equation*}
where $i=k,\dots,1$ and $\Delta_{k+1}=w_{k+1}=1$. Then 
\begin{equation*}
\nabla_{w_{i}}f=\Delta_{i}\,\sigma_{i-1}
\end{equation*}
where $\sigma_0=x$.

\mn
{\large\bf 5.  Gradient $\nabla_Wf(X;W)$ in general case}

\mn
In general case, the components $\nabla_{W_i}f(X;W)$ of the gradient
$\nabla_{W}f(X;W)$ can be written in a form, analogous to (1).
Recall that $W_i$ is $n_{i}\times n_{i-1}$ matrix and
hence $\nabla_{W_i}f(X;W)$ is also $n_{i}\times n_{i-1}$ matrix.
The application of the chain rule in such notation 
can not be called a quite simple problem. Nevertheless, using the analogy
with (1), one can guess the result.
To write down the formulas, we need three kind of matrix product:
the usual column-by-row product $A\cdot B$,
the Hadamard column-by-column product $A\circ B $
and the ``inverted'' column-by-matrix product
$A\bullet B=B\cdot A$ of the column $A$ by the matrix $B$.
As in the formulas (2), we omit, for brevity, the arguments $N_i$
in $\Sigma_i(N_i)$ and $\Sigma'_i(N_i)$.

\mn
{\it For the network function
\begin{equation*}
f(X;W)=\Sigma_k(W_k\cdot\Sigma_{k-1}(W_{k-1}\cdot\Sigma_{k-2}
\dots\cdot\Sigma_2(W_2\cdot\Sigma_1(W_1\cdot X))\dots))
\end{equation*}
we have 
\begin{equation}
\begin{cases}
\nabla_{W_k}f=\overbrace{\Sigma'_k\cdot\Sigma^{\,T}_{k-1}}\\
\nabla_{W_{k-1}}f=\underbrace{\Sigma'_k\bullet
W^{\,T}_k}\circ\overbrace{\Sigma_{k-1}'\cdot\Sigma^{\,T}_{k-2}}\\
\nabla_{W_{k-2}}f=\underbrace{\Sigma'_k\bullet
W^{\,T}_k}\circ\underbrace{\Sigma_{k-1}'\bullet W^{\,T}_{k-1}}
\circ\overbrace{\Sigma'_{k-2}\cdot\Sigma^{\,T}_{k-3}}\\
..........................................................................................\\
\nabla_{W_1}f=\underbrace{\Sigma'_k\bullet
W^{\,T}_k}\circ\underbrace{\Sigma_{k-1}'\bullet W^{\,T}_{k-1}}
\circ\underbrace{\Sigma'_{k-2}\bullet W_{k-2}}\,\dots\,
\underbrace{\Sigma_{2}'\bullet W^{\,T}_{2}}\circ\overbrace{\Sigma'_{1}\cdot X^{\,T}}\,.\\
\end{cases}
\end{equation}}

\mn
{\bf Remarks.}

\mn
{\bf 1.} The braces indicate the periodicity in the structure 
of formulas (rather than the order of multiplications). 
The actions order in (3) is from the left to the right.

\mn
{\bf 2.} One can replace the product $\Sigma'_k\bullet W^{\,T}_k$
by $\Sigma'_k\cdot W^{\,T}_k$,
(because $\Sigma'_k$ is a scalar) and the bullet is used in this case
only in order to emphasize the periodicity in the structure of formulas.

\mn
{\bf 3.} In order to write down the recursive formula, we set, omitting again 
the arguments in the notations,
\begin{equation*}
\Delta_{i}=\Delta_{i+1}\bullet W^T_{i+1}\circ\Sigma_{i}'=
(W^T_{i+1}\cdot\Delta_{i+1})\circ\Sigma_{i}'\,,
\end{equation*}
where $i=k,\dots,1$ and $\Delta_{k+1}=W^T_{k+1}=1$. Then
\begin{equation*}
\nabla_{W_{i}}f=\Delta_{i}\cdot\Sigma^T_{i-1}
\end{equation*}
where $\Sigma_0=X$.

\mn
{\bf 4.} $\Delta_i$ is a $n_i$-column and
$\Sigma^T_{i-1}$ is a $n_{i-1}$-string, thus $\nabla_{W_{i}}f$
is, as expected, a $n_{i}\times n_{i-1}$-matrix.

\mn
{\bf 5.} By reversing the order of factors, one can get rid of bullets
and rewrite (3) as
\begin{equation}
\begin{cases}
\nabla_{W_k}f=\overbrace{\Sigma^{\,T}_{k-1}\otimes\Sigma'_k}\\
\nabla_{W_{k-1}}f=\overbrace{\Sigma^{\,T}_{k-2}\otimes\Sigma'_{k-1}}\circ
\underbrace{W^{\,T}_k\cdot\Sigma'_k}\\
\nabla_{W_{k-2}}f=\overbrace{\Sigma^{\,T}_{k-3}\otimes\Sigma'_{k-2}}\circ
\underbrace{W^{\,T}_{k-1}\cdot\Sigma'_{k-1}}\circ
\underbrace{W^{\,T}_{k}\cdot\Sigma'_{k}}\\
..........................................................................................\\
\nabla_{W_1}f=\overbrace{X^{\,T}\otimes\Sigma'_{1}}\circ
\underbrace{W^{\,T}_{2}\cdot\Sigma'_{2}}\circ\dots
\underbrace{W^{\,T}_{k-2}\cdot\Sigma'_{k-2}}\circ
\underbrace{W^{\,T}_{k-1}\cdot\Sigma'_{k-1}}\circ
\underbrace{W^{\,T}_{k}\cdot\Sigma'_{k}}\,.\\
\end{cases}
\end{equation}
The actions order in (4) is from the right to the left, $\otimes$ is 
the Kronecker product, and, as before, 
the braces are not related to the order of operations.

\mn
{\bf 6.} We can also get rid of Hadamard product by replacing the 
columns $\Sigma'_{i}$ with square diagonal matrices $\widehat\Sigma'_{i}$
where $\Sigma'_{i}$ is the  diagonal. Then (4) turns into 
\begin{equation}
\begin{cases}
\nabla_{W_k}f=\Sigma^{\,T}_{k-1}\otimes \widehat\Sigma'_k\\
\nabla_{W_{k-1}}f=\Sigma^{\,T}_{k-2}\otimes(\widehat\Sigma'_{k-1}\cdot
W^{\,T}_k\cdot \widehat\Sigma'_k)\\
\nabla_{W_{k-2}}f=\Sigma^{\,T}_{k-3}\otimes(\widehat\Sigma'_{k-2}\cdot
W^{\,T}_{k-1}\cdot \widehat\Sigma'_{k-1}\cdot
W^{\,T}_{k}\cdot \widehat\Sigma'_{k})\\
..........................................................................................\\
\nabla_{W_1}f=X^{\,T}\otimes(\widehat\Sigma'_{1}\cdot
W^{\,T}_{2}\cdot \widehat\Sigma'_{2}\cdot\dots
W^{\,T}_{k-2}\cdot \widehat\Sigma'_{k-2}\cdot
W^{\,T}_{k-1}\cdot \widehat\Sigma'_{k-1}\cdot
W^{\,T}_{k}\cdot \widehat\Sigma'_{k})\,.\\
\end{cases}
\end{equation}
Note that $W^T_k$ is a column and $\widehat\Sigma'_{k}=\Sigma'_{k}$ is a scalar.

\mn
{\large\bf 6. Proof of (3)}

\mn
The formulas (3) follow immediately from the equalities
\begin{equation}
\nabla_{W_r}f=(\nabla_{\Sigma_r}f)\circ\Sigma_r'\cdot\Sigma^T_{r-1}
\end{equation}
\begin{equation}
\nabla_{\Sigma_r}f=(\nabla_{\Sigma_{r+1}}f)\circ\Sigma_{r+1}'\bullet\,W^T_{r+1}\,.
\end{equation}
and we will prove (6) and (7).
We need the following rule for calculating the gradient, 
consistent with our agreement on the matrix representation 
of matrix derivatives: if $A$ is a string and $B$ is a column, then
\begin{equation*}
\nabla_A(A\cdot B)=B^T\,.
\end{equation*}
Denote by $W^i_{r}$ the $i$-th raw of the matrix $W_{r}=\{W^i_{r}\}_i$
and by $\Sigma_{r}^i$ the $i$-th element of the column $\Sigma_{r}=\{\Sigma^i_{r}\}_i$.
In particular, $\nabla_{W_{r}}=\{\nabla_{W^i_{r}}\}_i$ and
$\nabla_{\Sigma_{r}}=\{\nabla_{\Sigma^i_{r}}\}_i$.
Let us verify now (6):
\begin{equation*}
\nabla_{W_r}f=\{\nabla_{W^i_r}f\}_i=
\{\nabla_{\Sigma^i_r}f\cdot\nabla_{W^i_r}\Sigma^i_{r}\}_i=
\{\nabla_{\Sigma^i_r}f\cdot\nabla_{W^i_r}\,(\Sigma^i_{r}\,(W^i_r\cdot\Sigma_{r-1}))\}_i=
\end{equation*}
\begin{equation*}
=\{\nabla_{\Sigma^i_r}f\cdot{\Sigma^i_{r}}'\cdot\Sigma_{r-1}^T\}_i=
(\nabla_{\Sigma_r}f)\circ\Sigma_r'\cdot\Sigma^T_{r-1}\,.
\end{equation*}
Next, the equation (7) in coordinate-wise form is
\begin{equation*}
\{\nabla_{\Sigma^i_r}f\}_i=
\{(\nabla_{\Sigma_{r+1}}f)\circ\Sigma_{r+1}'\bullet\,(W^T_{r+1})^i\}_i\,.
\end{equation*}
Let us verify the equality of the corresponding coordinates:
\begin{equation*}
\nabla_{\Sigma^i_r}f=
\langle\nabla_{\Sigma_{r+1}}f\,,\nabla_{\Sigma_r^i}\Sigma_{r+1}\rangle=
\langle\nabla_{\Sigma_{r+1}}f\,,\nabla_{\Sigma_r^i}(\Sigma_{r+1}(W_{r+1}\Sigma_r))\rangle=
\end{equation*}
\begin{equation*}
=\langle\nabla_{\Sigma_{r+1}}f\,,\Sigma_{r+1}'\circ (W_{r+1}^T)^i\rangle
=\langle\nabla_{\Sigma_{r+1}}f\circ\Sigma_{r+1}'\,, (W_{r+1}^T)^i\rangle=
(\nabla_{\Sigma_{r+1}}f)\circ\Sigma_{r+1}'\bullet\,(W^T_{r+1})^i\,
\end{equation*}
(here $\langle\,\,,\,\rangle$ is the scalar product).

\end{document}